\documentclass{article}

\usepackage{arxiv}

\usepackage[utf8]{inputenc} 
\usepackage[T1]{fontenc}    
\usepackage{hyperref}       
\usepackage{url}            
\usepackage{booktabs}       
\usepackage{amsfonts}       
\usepackage{nicefrac}       
\usepackage{microtype}      
\usepackage{amsmath}
\DeclareMathOperator*{\argmax}{arg\,max}

\usepackage{cleveref}       
\usepackage{lipsum}         
\usepackage{graphicx}
\usepackage{natbib}
\usepackage{doi}

\usepackage{graphicx}
\graphicspath{ {Figures} }

\title{Target Variable Engineering}

\newif\ifuniqueAffiliation
\uniqueAffiliationtrue

\ifuniqueAffiliation 
\author{Jessica Clark\thanks{\url{http://www.jessicamarieclark.com}}\\
	Decisions, Operations, and Information Technology Department\\
	Robert H. Smith School of Business \\
	University of Maryland - College Park\\
	College Park, MD 20912\\
	\texttt{jmclark@umd.edu} \\
}

\renewcommand{\headeright}{ }
\renewcommand{\undertitle}{ }
\renewcommand{\shorttitle}{Target Variable Engineering}

\hypersetup{
pdftitle={Target Variable Engineering},
pdfauthor={Jessica M.~Clark},
pdfkeywords={Machine Learning, Target Variables, Hyperparameter Optimization},
}

\begin{document}
\maketitle

\begin{abstract}
How does the formulation of a target variable affect performance within the ML pipeline? The experiments in this study examine numeric targets that have been binarized by comparing against a threshold. We compare the predictive performance of regression models trained to predict the numeric targets vs. classifiers trained to predict their binarized counterparts. Specifically, we make this comparison at every point of a randomized hyperparameter optimization search to understand the effect of computational resource budget on the tradeoff between the two. We find that regression requires significantly more computational effort to converge upon the optimal performance, and is more sensitive to both randomness and heuristic choices in the training process. Although classification can and does benefit from systematic hyperparameter tuning and model selection, the improvements are much less than for regression. This work comprises the first systematic comparison of regression and classification within the framework of computational resource requirements. Our findings contribute to calls for greater replicability and efficiency within the ML pipeline for the sake of building more sustainable and robust AI systems.
\end{abstract}

\keywords{Machine Learning \and Target Variables \and Hyperparameter Optimization}

\section{Introduction}

Target variables for machine learning applications should be formulated to support a specific decision, and in research contexts are usually treated as a fixed part of the ML pipeline. However, even given a specific task, there can be flexibility in the formulation of the target variable. Specifically, there are many applications where either numeric or categorical predictions could be equally suitable. To use a classic example, if a company wants to address customers who are likely to ``churn", i.e. suspend their services in the near future, they could train regression models to predict each customer's numeric future usage of their service. Alternatively, the company could binarize the target variable based on whether usage is above or below some threshold. Then, they would train binary classifiers to predict the resulting categorical target. Beyond the obvious differences between the two formulations, such as choosing the appropriate evaluation metric, the choice between these two potential target variable formulations is not usually discussed in the extant ML literature.

This work studies the fundamental but previously unanswered research question of how regression vs. classification models differ, in terms of both resource requirements within the ML pipeline and replicability of results. In applications where either formulation could be used interchangeably, the choice is usually approached heuristically. To systematize this choice, we conduct an experimental comparison of various parts of the ML pipeline given a numeric target versus the categorical target variable that results from binarizing using a threshold. Thus, the predictive problems compared are identical, save for the formulation of the target variable. 

A key component of the experiments is that we compare the performance for the two task types as a function of Hyperparameter Optimization (HPO) random search budgets. Across feature sets, target variables, and model families, we consistently find that regression tasks require significantly more computation to converge on optimal parameters than their classification counterparts. Digging into these results reveals that regression is more sensitive not only to HPO budget, but to all of the heuristic choices across the ML pipeline that we investigate. HPO budget, model selection, choice of grid search algorithm, and amount of training data all yield significantly more variation in terms of test performance than they do for classifiers. The performance of regression models are also more sensitive to randomness and therefore prone to overfitting.

Thus, in applications where either formulation could be appropriate, choosing classification enables use of smaller HPO budgets and yields more straightforward generalization of results. In general, modelers planning regressions should ensure a large budget for HPO and use repeated sampling to ensure generalizability. Modelers conducting classification don't need to use such large grids as are used for regression, and can also use smaller amounts of training data to reach nearly optimal results. Given the substantial carbon emissions associated with HPO, recent work has pointed out that prioritizing computationally efficient algorithms can lead to significant reductions in environmental impact \citep{strubell2019energy, schwartz2020green}. Our findings thus contribute to research in sustainable AI not by developing more efficient algorithms, but by streamlining other parts of the ML pipeline. Our findings also contribute to advances in automated machine learning (AutoML) by systematizing some heuristic choices. Finally, we also contribute to work dealing with the crisis of replicability in science broadly, and ML more specifically \citep{bouthillier2021accounting}. Simply put, classification results are easier to replicate. They are less susceptible to overfitting, less sensitive to both randomness and heuristic choices. Although the state-of-the-art is relatively better (on average) for regression, making decisions systematically and reporting all parameters is of more critical importance for this task.

\section{Related Work}

Practical guides for applied machine learning emphasize the importance of formulating the target variable to align with some decision that is being supported \citep{provost2013data}. For instance, CRISP-DM, a widely-used business framework for applying machine learning, includes formulating the target variable as part of the ``business understanding" phase \citep{chapman2000crisp}. It has been acknowledged that tasks related to the business understanding phase are not widely studied in the literature \citep{baier2019challenges}, and most research in ML (both applied and theoretical) assumes that the target variable is a fixed concept.

There are a few common practices relating to modifying the target variable to make prediction problems easier. For example, if the distribution of numeric values has a heavy right tail, it can be log-transformed. If a binary-valued target variable has a strong class imbalance, oversampling or undersampling can be used to improve predictive performance. Target variables can be specially formulated for particular applications, such as causal effect estimation \citep{fernandez2022causal}. The field of \emph{prompt engineering} includes constructing tasks to elicit the best-possible classifications or predictions from large language models such as ChatGPT (i.e. \citet{sorensen2022information}, \citet{brown2020language}, \citet{liu2021pretrain}, \citet{zhou2022large}).

There are many examples of past work which have implicitly compared classification and regression for a particular task by providing reasons for binarizing numerical target variables. First, it may be easier to acquire binary, rather than numerical, labels, especially when the labels are user-generated \citep{sparling2011rating}. Second, although what is being measured directly may be numeric, typical use of that variable involves a categorical decision \citep{liu2020predicting}. Binarization may result in a simpler problem \citep{zhang2021measuring} or yield desirable evaluation metrics such as a confusion matrix \citep{abbasi2019don}. Recent work has also studied how and why reformulating a regression problem as classification can result in improved performance of neural networks \citep{stewart2023regression}, which they term ``the binning phenomenon." 

The question of whether it's ever appropriate to binarize a numeric dependent variable has also been debated in the traditional statistics literature, and is generally viewed as a bad practice \citep{royston2006dichotomizing, fitzsimons2008death}. Binarizing has been found to lead to misleading results in the size and direction of coefficients in regression analysis \citep{maxwell1993bivariate}. Although binarizing the response variable makes results easier to explain and present to non-practitioners, it can also lead to a loss of information and statistical power \citep{irwin2003negative}. The field has continued to discuss the role of dichotomization in statistics \citep{pham2015ok}. The results in this work do not directly contradict past findings; however, we find that there are positive benefits to binarizing in predictive contexts.

\section{EXPERIMENTS}

The core experiments in this paper seek to compare the process and performance models trained to predict numerical target variables (regression task) versus binary categorical target variables (classification task). The experimental framework relies on the idea that we can compute a binarized counterpart to any numeric target variable by comparing to a threshold. Thus, all of the other parameters of the experiment are kept as similar as possible such that the only difference is the two target variable data types.

\subsection{Grid Search}

Following the notation developed by \cite{dodge2019show}, we denote $\mathcal{M}$ to indicate the \emph{model family}, meaning a general induction algorithm with a set of $k$ hyperparmeters that can be optimized. Each $k$-tuple of values of individual hyperparameters forms one \emph{hyperparameter value} $h$, and the set of all possible hyperparameter values forms $\mathcal{H}_{\mathcal{M}}$. In our experiments, we choose model families that can be adapted to predict either numeric or categorical target variables, and thus $\mathcal{H}_{\mathcal{M}}$ is the same for both task types.

The grid searches completed in this paper conduct $B$ random draws from $\mathcal{H}_{\mathcal{M}}$, and are randomly initialized $S$ times. Let $\mathcal{A}\left(\mathcal{M}, h, s, \mathcal{D}_{T}, \mathcal{D}_{P} \right)$ denote an algorithm that returns the performance in some prediction data $\mathcal{D}_{P}$ using a model from $\mathcal{M}$ with hyperparameter value $h$ trained on $\mathcal{D}^{T}$, given random initialization state $s \in \{1,\cdots, S \}$. For draw $b$ from $\mathcal{H}_{\mathcal{M}}$, define the validation and test performance for draw $b$ as:

\begin{align}
v_b &= \mathcal{A}\left(\mathcal{M}, h_b, s, \mathcal{D}_{T}, \mathcal{D}_{V} \right) \\
t_b &= \mathcal{A}\left(\mathcal{M}, h_b, s, \mathcal{D}_{T}, \mathcal{D}_{TE} \right)
\end{align}

We report the cumulative maximum validation performance after $B$ grid search iterations $v^*_B$, the best hyperparameter value $h^*_B$ and test performance using those best hyperparameters $t^*_B$:

\begin{align}
v^*_B &= \max_{h \in \{h_1,...,h_B\} } \mathcal{A}\left(\mathcal{M}, h, s, \mathcal{D}_{T}, \mathcal{D}_{V} \right) \\
h^*_B &= \argmax_{h \in \{h_1,...,h_B\} } \mathcal{A}\left(\mathcal{M}, h, s, \mathcal{D}_{T}, \mathcal{D}_{V} \right) \\
t^*_B &= \mathcal{A}\left(\mathcal{M}, h^*_B, s, \mathcal{D}_{T}, \mathcal{D}_{TE} \right)
\end{align}

For the experiments in this paper, we set $B=400$ and $S = 15$. The first draw for each search always comprises the default hyperparameters for $\mathcal{M}$, yielding a reasonable estimate of off-the-shelf performance. $\mathcal{H}_{\mathcal{M}}$ (including the default parameters) is specific to each $\mathcal{M}$ and were drawn from Hyperopt-Sklearn \citep{komer2014hyperopt} and have been used in past work on HPO \citep{grinsztajn2022tree, gorishniy2021revisiting}. 

We experiment with both a standard random search \citep{bergstra2012random} as well as the Tree-Structured Parzen Estimator algorithm, a Bayesian optimization algorithm \citep{turner2021bayesian}. We use the Optuna library in Python for managing this grid search \citep{akiba2019optuna}.

\subsection{Datasets}

The data testbed uses three feature sets gathered from publicly-available online data: Airbnb.com\footnote{http://insideairbnb.com/get-the-data/, accessed March 2018}, Kickstarter.com \footnote{https://webrobots.io/kickstarter-datasets/, accessed Dec 2015}, and Yelp.com \footnote{https://www.yelp.com/dataset, accessed Jan 2022}. We engineered a tabular feature set of size approximately 2000 from each.\footnote{Note that other tabular benchmarking datasets \citep{grinsztajn2022tree} mostly have considerably fewer features; having larger feature sets allows us to experiment with feature set size.}

We derived 10 numeric target variables from each domain, which were standardized using $z$-score normalization such that each one has mean 0 and standard deviation 1. We further created a binarized counterpart to each numeric target by thresholding at the mean value. That is, the binarized target is positive if the numeric target is greater than 0, and negative otherwise. Table \ref{table:data} contains detailed descriptions of the datasets and target variables.

\begin{table}[]
\centering
\begin{tabular}{|p{.15\linewidth}|p{.3\linewidth}|p{.55\linewidth}|}
\hline
\textbf{Domain} & \textbf{Feature Set Description}                                                       & \textbf{Numeric Target Variables}                                                                                                                                                                                                                                                                                                                                                  \\ \hline
Airbnb               & Information and descriptions of listings from Airbnb.com.                              & (1) Number of guests accommodated (2) Availability in the next 30 days (3) Availability in the next 60 days (4) Availability in the next 90 days (5) Availability in the next 365 days (6) Host listings count (7) Number of reviews (8) Price (9) Average rating (10) Average reviews per month                                                                                   \\ \hline
Kickstarter          & Information and descriptions of completed crowdfunding campaigns from Kickstarter.com. & (1) Dollars pledged (2) Number of backers (3) Dollar goal amount (4) Number of reward levels for contributors (5) Minimum amount to receive an award (6) Maximum amount to receive an award (7) Standard deviation of reward amounts (8) Time between campaign creation and campaign launch (9) Number of sentences in description (10) Average length of sentences in description \\ \hline
Yelp                 & Information about business which have received reviews on Yelp.com.                    & (1) Total number of reviews (2) Average star rating (3) Average "useful" review rating (4) Average "funny" review rating (5) Average "cool" review rating (6) Average review count of reviewers (7) Percent of reviewers with "elite" status (8) Percent of male reviewers (9) Number of checkins (10) Number of tips                                                              \\ \hline
\end{tabular}
\caption{Description of feature sets and numeric target variables. \label{table:data}}
\end{table}

For consistency of comparison, each feature set contains $30,000$ instances, yielding ``medium"-sized data.\footnote{We have experimented with much larger datasets in terms of both instance and feature set sizes and found very consistent results but due to computational resource constraints we have excluded a full comparison from this paper.} Most results in the paper, other than those presented in Section \ref{sect:learningcurves}, divide each feature set into three:

\begin{enumerate}
\item $\mathcal{D}^{T}$: $10,000$ training instances.
\item $\mathcal{D}^{V}$: $5,000$ validation instances, used for tuning hyperparameters.
\item $\mathcal{D}^{TE}$: $15,000$ test instances, used for evaluation.
\end{enumerate}

\subsection{Model Families}

The experiments in this paper use three families of induction algorithms that can be suitable for either the regression or classification task. First, ensemble methods such as XGBoost \citep{chen2015xgboost} are currently regarded as the state-of-the-art ML model for tabular data \citep{grinsztajn2022tree, borisov2022deep, shwartz2022tabular} and so most of our main results use XGBoost for modeling. Second, although ensemble methods currently have superior performance, deep learning for tabular data is an area of active research, and a recent survey found that ResNet and other deep learning models can achieve comparable or superior performance to XGBoost on benchmarking datasets, although they generally take far longer to train \citep{gorishniy2021revisiting}. Third, we include $L_2$-regularized linear methods (linear regression and logistic regression) as a third model family because they are simple to understand and interpret, are in common use across a wide variety of fields and applications, and have been found to achieve decent performance in past work \citep{rudin2019stop, clark2019unsupervised}.

\subsection{Evaluation}

We measure the $R^2$ regression score between the actual numeric values and numerical predictions. $R^2$ is normally between 0 and 1. Our results include numerous modeling settings yielding negative $R^2$ due to overfitting to the training data. To ensure a fair comparison for such results, we truncated the reported $R^2$ to 0. For the corresponding classifiers, we measured the AUC (Area under the ROC Curve), which represents the ability of a classifier's scores to rank positive instances above negative ones \citep{provost2001robust} and is usually between 0.5 and 1. AUCs of less than 0.5 in the validation or test data were truncated to 0.5.

A direct ``apples-to-apples" comparison of regression and classification results is challenging for two reasons. First, the two performance measures are on different scales and measure different things.  Second, even among target variables of the same type, performance is not necessarily comparable; some tasks are easier and some are harder. Therefore, we normalize both $R^2$ and AUC relative to the maximum value achieved in each random initialization of grid search. That is, we compare progress from the minimum possible value to the maximum value as a function of the HPO budget. If $v_{min}$ is the minimum achievable value (0 for $R^2$ and $0.5$ for AUC), then define:

\begin{align}
vnorm_b &= \frac{v_b - v_{min}}{v^*_B - v_{min}} \label{eq:vnorm} \\
tnorm_b &= \frac{t_b - v_{min}}{v^*_B - v_{min}} \label{eq:tnorm}
\end{align}

Thus, $vnorm_b$ starts somewhere between 0 and 1 with the default $h$ value. As $b$ increases to $B$, $vnorm_b$ increases to 1. We expect that $tnorm(b,s)$ is less than $vnorm(b,s)$, and shows the relative generalizability of each random search run by comparing the test performance to the expected maximum (validation) performance.

\section{RESULTS}

The experiments in this paper illustrate key differences in how the model selection and training process plays out for two types of predictive tasks: regression (predicting numerical targets) vs. classification (predicting their binarized counterparts). In summary, this section shows that regression requires more time and data resources to reach optimal performance, and is also more sensitive to various settings in the process.

Unless stated otherwise, most of the results in this section use feature set sizes of approximately 2000, a random sampling algorithm for HPO, and XGBoost as the model family. Section \ref{sect:pipeline} probes the effect of these three choices.

\subsection{Hyperparameter Optimization}

\begin{figure}
\centering
\begin{minipage}[t]{.45\textwidth}
 \begin{flushleft}
  \includegraphics[width=\linewidth]{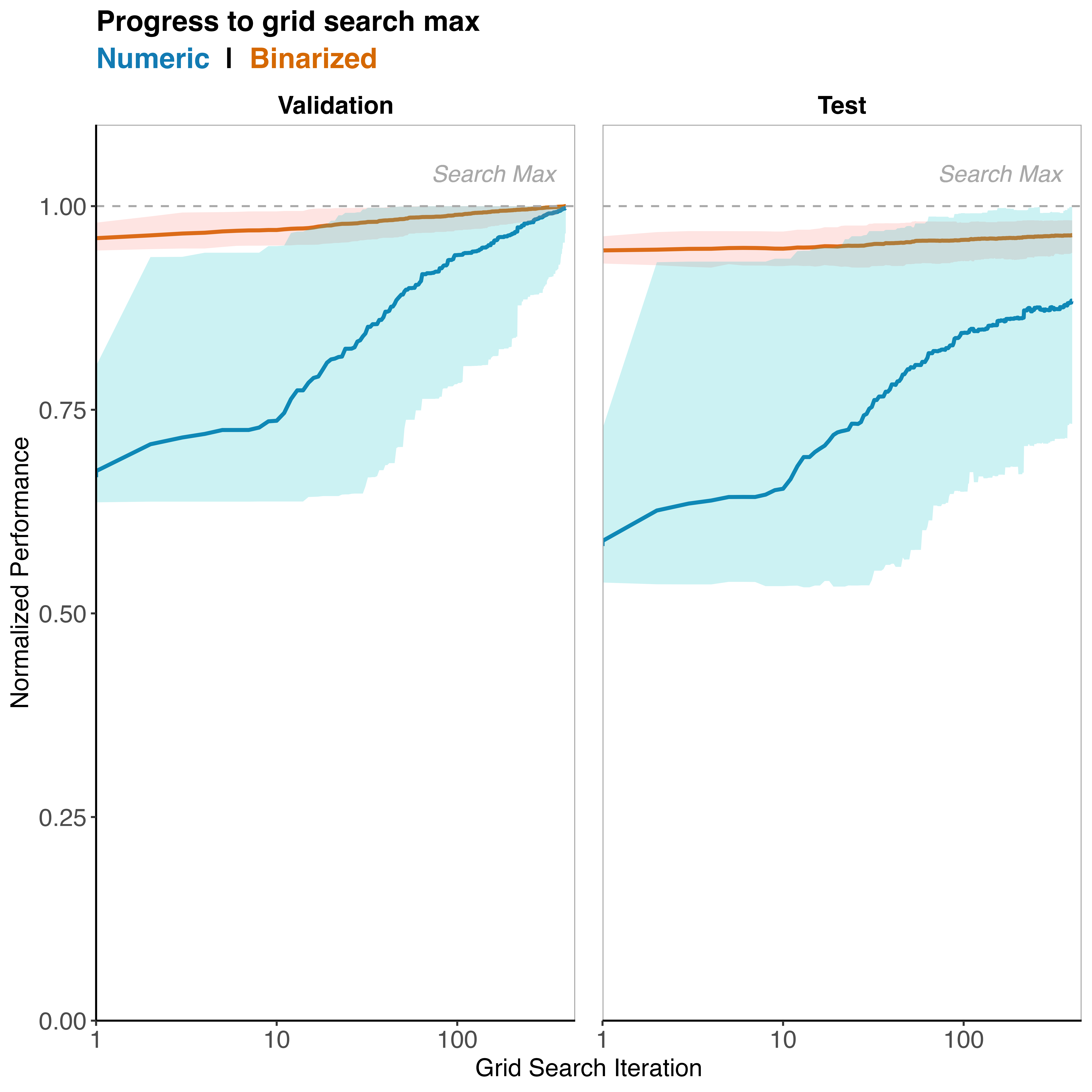}
\caption{Cumulative maximum validation and test performance by HPO budget for numeric and binarized target variables, normalized relative to the maximum overall validation performance. \label{fig:binvnum}}
  \end{flushleft}
\end{minipage}%
\hfill
\begin{minipage}[t]{.45\textwidth}
  \begin{flushright}
	\includegraphics[width=\linewidth]{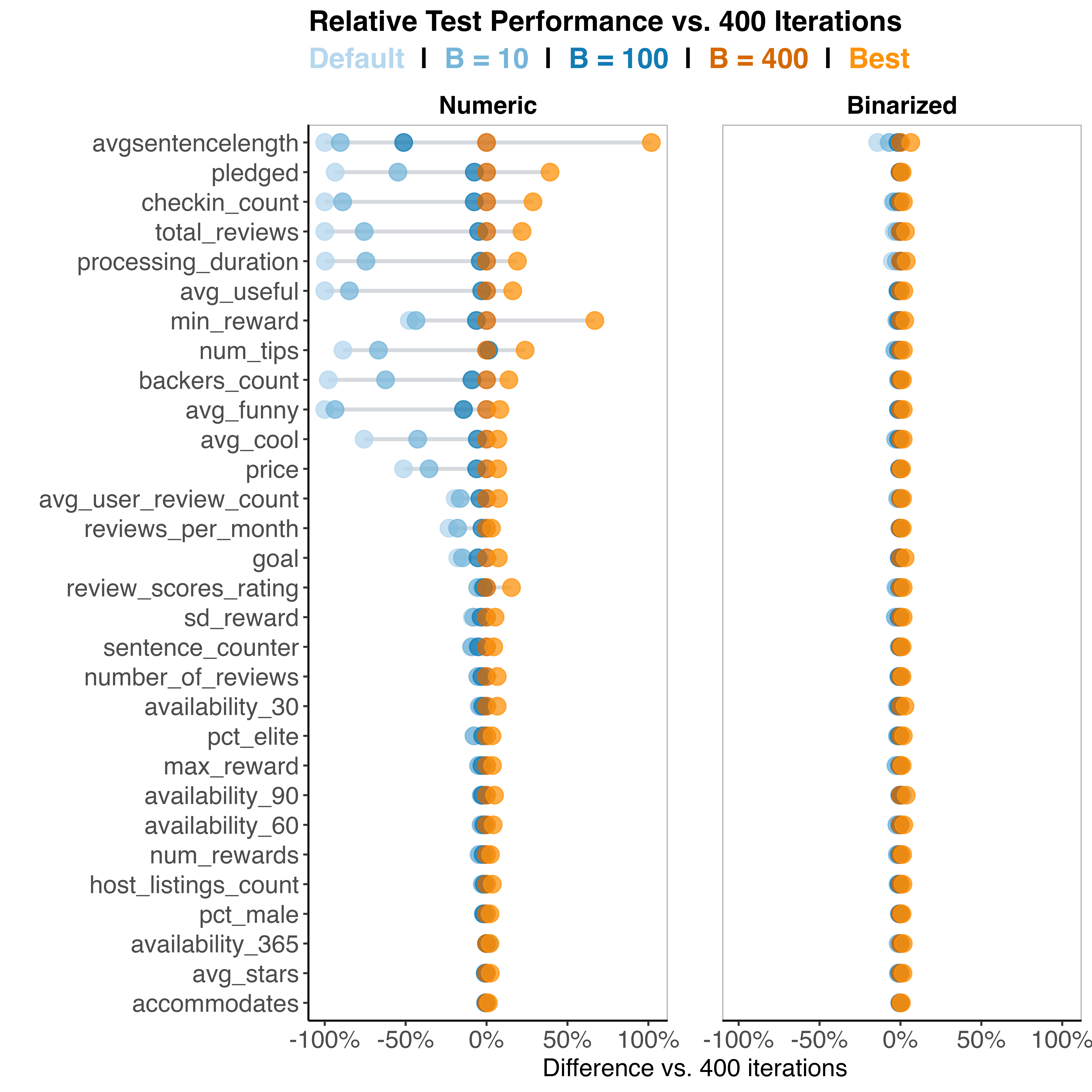}    
	\caption{Average $t^*$ given various budgets, relative to a budget of 400 iterations, also including maximum performance across all random seeds.\label{fig:tuning}}
    \end{flushright}
\end{minipage}
\end{figure}

Using the formulas given in Equations \ref{eq:vnorm} and \ref{eq:tnorm}, Figure \ref{fig:binvnum} plots the normalized cumulative maximum validation and test performance for each numeric target variable vs. its binarized counterpart across 400 HPO budgets. The lines show the average performance across 30 target variables and 15 random initializations, and the shaded regions show the average difference across target variables between the minimum and maximum initializations. The validation performance of regression tasks (in blue) not only has relatively worse performance given default $h$, but also requires a higher budget to approach $v^*$ . Table \ref{tab:grid_convergence} summarizes the average number of trials required to reach $90$, $95$, and $100\%$ of $v^*$ for each target variable and random initialiation. The differences between numeric and binarized targets are all significant for $\alpha > .99$.

%

\begin{table}
\caption{Trials Required for Grid Search Convergence \label{tab:grid_convergence}}
\centering
\begin{tabular}[t]{l|l|l|l|l|l}
\textbf{$\%$ of Max} & \textbf{Num} & \textbf{Bin} & \textbf{Diff} & \textbf{Std Err} & \textbf{p-val}\\
\hline
90\% & 48.19 & 0.60 & 47.59 & 3.85 & $< .001$\\
95\% & 86.92 & 6.41 & 80.51 & 4.92 & $< .001$\\
99\% & 162.94 & 110.21 & 52.73 & 7.49 & $< .001$\\
\end{tabular}
\end{table}

Furthermore, the HPO process for regression is less generalizable. Note in Figure \ref{fig:binvnum} that the test performance for regression is lower than for classification, relative to what would be expected given the validation performance. After 400 search iterations, the average regression $t^*$ is $0.88$. For classification, it is $0.96$. A t-test for the difference between these two means is significant for $\alpha = 0.99$. The average gap between the minimum and maximum $tnorm*$ is also larger for regression, so the test performance has greater variation relative to the expected validation performance. After 400 iterations, the average $tnorm*$ range is 0.27 for regression and 0.04 for classification. Again, a $t$-test for the difference in these two means is significant for $\alpha = 0.99$. These results suggests that regression would benefit from increased HPO budgets, i.e. more computational resources. Certainly, classification and regression should not use the same sizes of grid.
 
These results are further emphasized in Figure \ref{fig:tuning}, which compares the average $t^*$ given a budget of 400 iterations versus various other budgets: the default HP, 10, and 100 iterations as well as the overall best $h$ found across all random initializations for each target variable. For many numeric variables, there is a substantial loss in performance for the default $h$ and other smaller budgets. Furthermore, even the average $t^*$ after 400 iterations is still far from the true best possible performance across random searches, again emphasizing the tendency of regression to overfit. These differences are not present for classification. Table \ref{tab:tuning_diffs} summarizes the average percent difference versus a budget of 400 iterations for numeric and binarized targets. All differences between numeric and binarized tasks are significant for $\alpha = 0.99$ based on paired t-tests. Not tuning, or using a smaller grid, affects classification significantly less than it affects regression. Also, the best possible outcome is substantially larger for regression than the average, again calling the replicability and generalizability of regression results into question.

%

\begin{table}
\caption{Mean Difference vs. 400 Tuning Iterations \label{tab:tuning_diffs}}
\centering
\begin{tabular}[t]{l|l|l}
\textbf{Iterations} & \textbf{Num (std)} & \textbf{Bin (std)}\\
\hline
0 & -39.12$\%$ (0.425) & -0.56$\%$ (0.004)\\
10 & -30.64$\%$ (0.333) & -0.51$\%$ (0.003)\\
100 & -5.33$\%$ (0.091) & -0.19$\%$ (0.001)\\
Best Overall & 13.91$\%$ (0.199) & 0.58$\%$ (0.002)\\
\end{tabular}
\end{table}

\subsection{Learning Curves} \label{sect:learningcurves}

The results in this section used the process given by \citet{perlich2003tree} to create learning curves that show generalization performance with respect to the amount of training data for regression vs. classification. In order to experiment with larger quantities of training data (up to $20,000$ training instances), we recombined $\mathcal{D}^{T}$, $\mathcal{D}^{V}$, and $\mathcal{D}^{TE}$, then randomly selected $5,000$ test instances for each target. To create learning curves, we repeated the following steps 30 times. 

\begin{enumerate}
\item Randomly draw $k$ training instances, where $k$ is between $100$ and $20,000$.
\item Using the training set of size $k$, train an XGB model using a best $h$ to predict the numeric target. Estimate predictions in the test set and measure the $R^2$.
\item Using the training set of size $k$, train an XGB model using a best $h$ to predict the binarized target. Estimate predictions in the test set and measure the AUC.
\item Normalize each $R^2$ and $AUC$ such that 0 is the minimum possible performance and 1 is the maximum observed performance across all $k$ for that target.
\end{enumerate}

\begin{figure}
\centering
\begin{minipage}[t]{.45\textwidth}
 \begin{flushleft}
  \includegraphics[width=\linewidth]{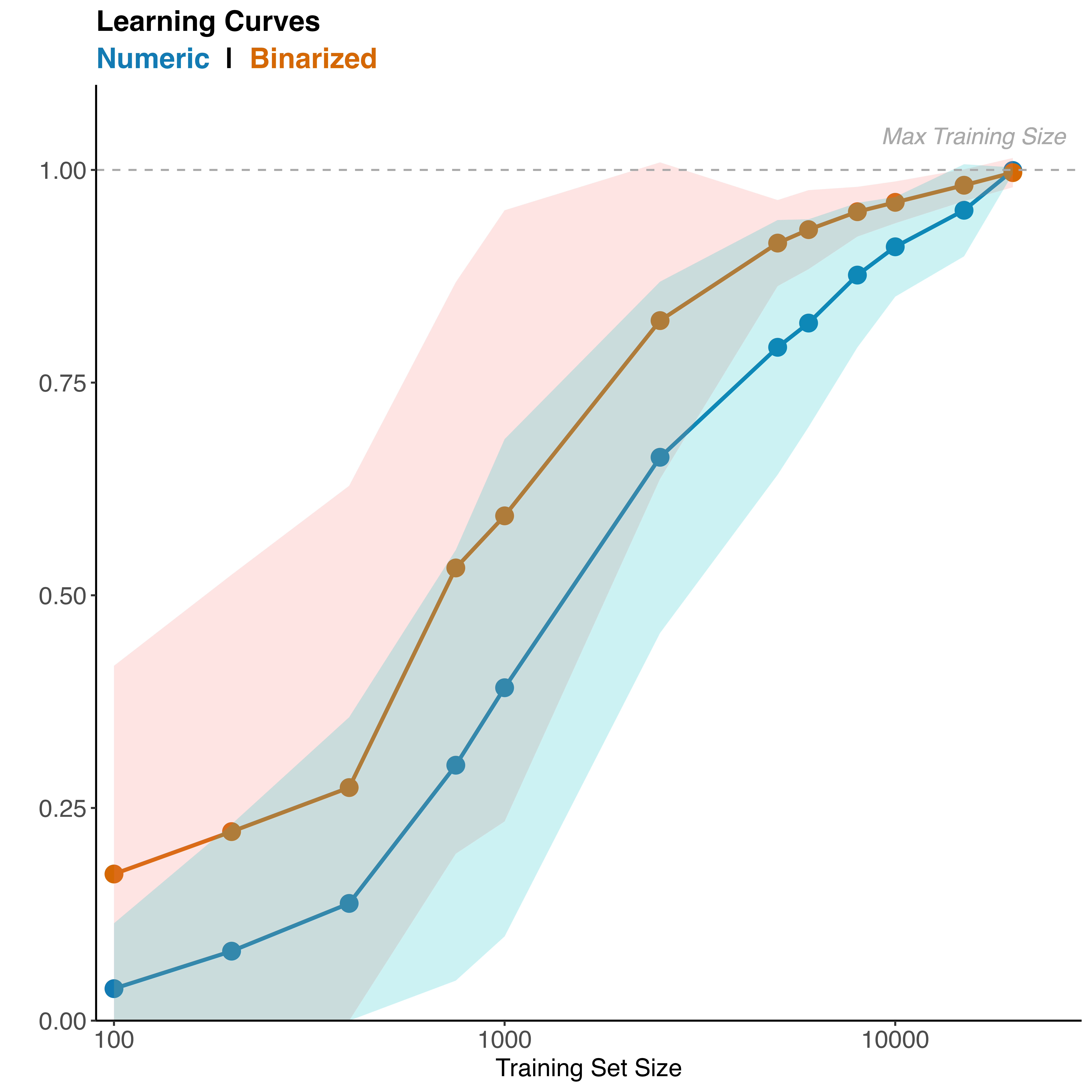}
\caption{Average normalized learning curves for regression vs. classification. On average, regression has steeper learning curves, meaning that the performance is more sensitive to training set size.\label{fig:learning_curve}}
  \end{flushleft}
\end{minipage}%
\hfill
\begin{minipage}[t]{.45\textwidth}
  \begin{flushright}
	\includegraphics[width=\linewidth]{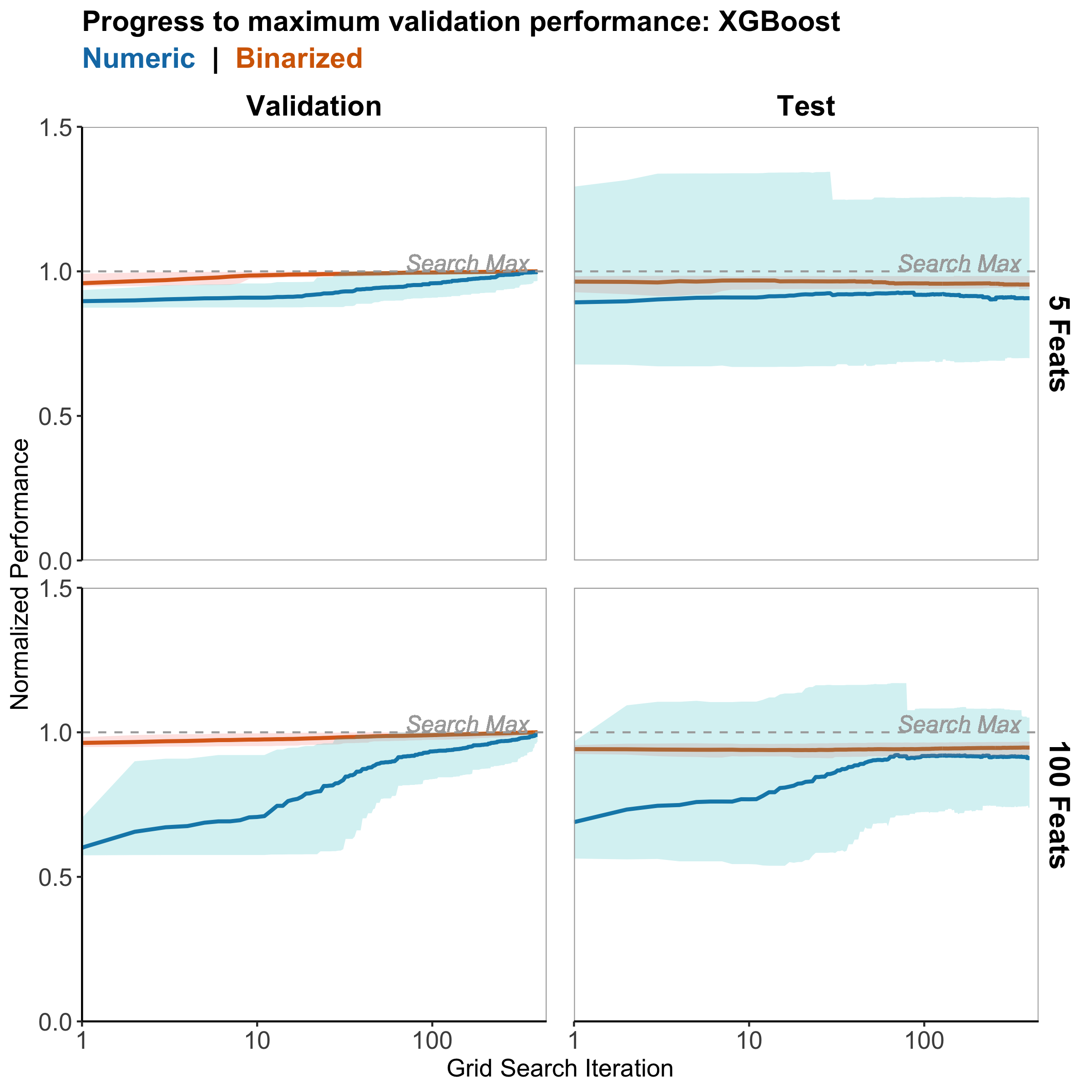}    
	\caption{Validation and test performance across feature set sizes. \label{fig:datasize}}
    \end{flushright}
\end{minipage}
\end{figure}

Figure \ref{fig:learning_curve} shows the normalized progress to the maximum performance averaged across 30 target variables and 30 random draws for each. The shaded regions represent a +/- 1 standard deviation interval around the average. This chart provides evidence that the learning curves for regression are ``steeper" with respect to the amount of training data. That is, for any number of training data, classification tends to have relatively closer performance to the maximum than regression. For instance, the average normalized performance for regression with a 100-instance training set is $4\%$ of the maximum observed and $17\%$ for classification. With a 1000-instance training set, regression is at $39\%$ and classification is at $59\%$.

We also expect that learning curves will level out as the marginal benefit of more data diminishes. At the high end of training set sizes, the classification learning curves appear to be flattening, while the regression curves are apparently still increasing. This implies that regression models receive relatively more benefit from more data; once again, classification requires fewer resources to perform at the highest level.

%

\subsection{Other parts of the pipeline} \label{sect:pipeline}

There are other heuristic choices in the ML pipeline besides HPO budget. This section probes the effects of the size of the feature set, the choice of sampling algorithm in the HPO grid search, and the choice of the model family for regression vs. classification.

\subsubsection{Feature set size}
The models trained in the prior sections were trained using approximately 2000 features each. What happens if there are fewer features? Figure \ref{fig:datasize} replicates Figure \ref{fig:binvnum} with 5 and 100 features. With 5 features, the relative differences between regression and classification are less dramatic, although still present. The differences are quite evident when there are 100 features. Anecdotally, we note that we have conducted preliminary experiments both on published benchmark tabular datasets \citep{grinsztajn2022tree} as well as feature sets with up to $200,000$ features, and the results of these preliminary experiments confirm the main results in this paper.


\subsubsection{Grid Search Sampling Algorithm} \label{sect:sampler}

\begin{figure}
\centering
\begin{minipage}[t]{.45\textwidth}
 \begin{flushleft}
  \includegraphics[width=\linewidth]{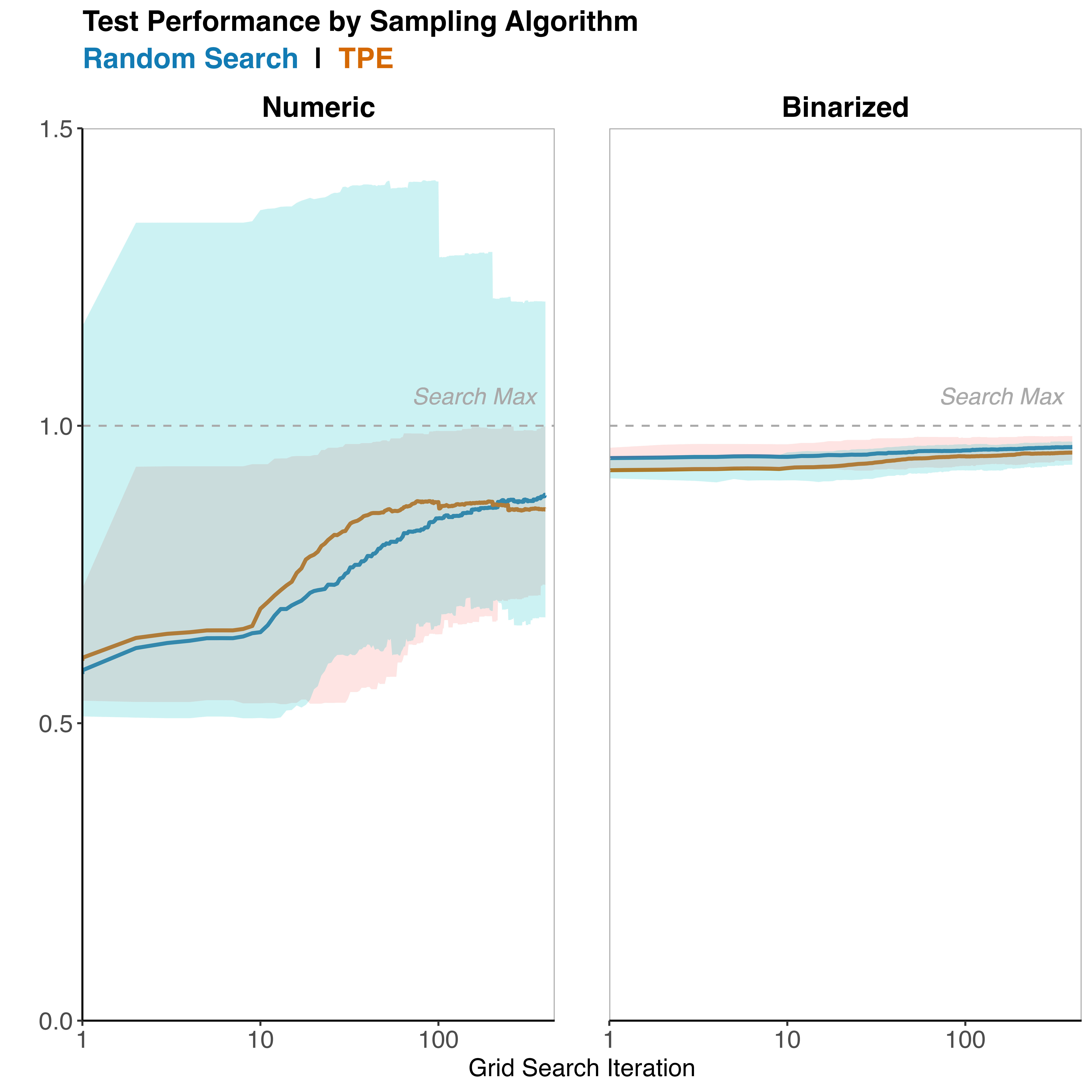}
\caption{Relative performance of random sampling vs. Tree Parzen Estimator. \label{fig:sampler}}
  \end{flushleft}
\end{minipage}%
\hfill
\begin{minipage}[t]{.45\textwidth}
  \begin{flushright}
	\includegraphics[width=\linewidth]{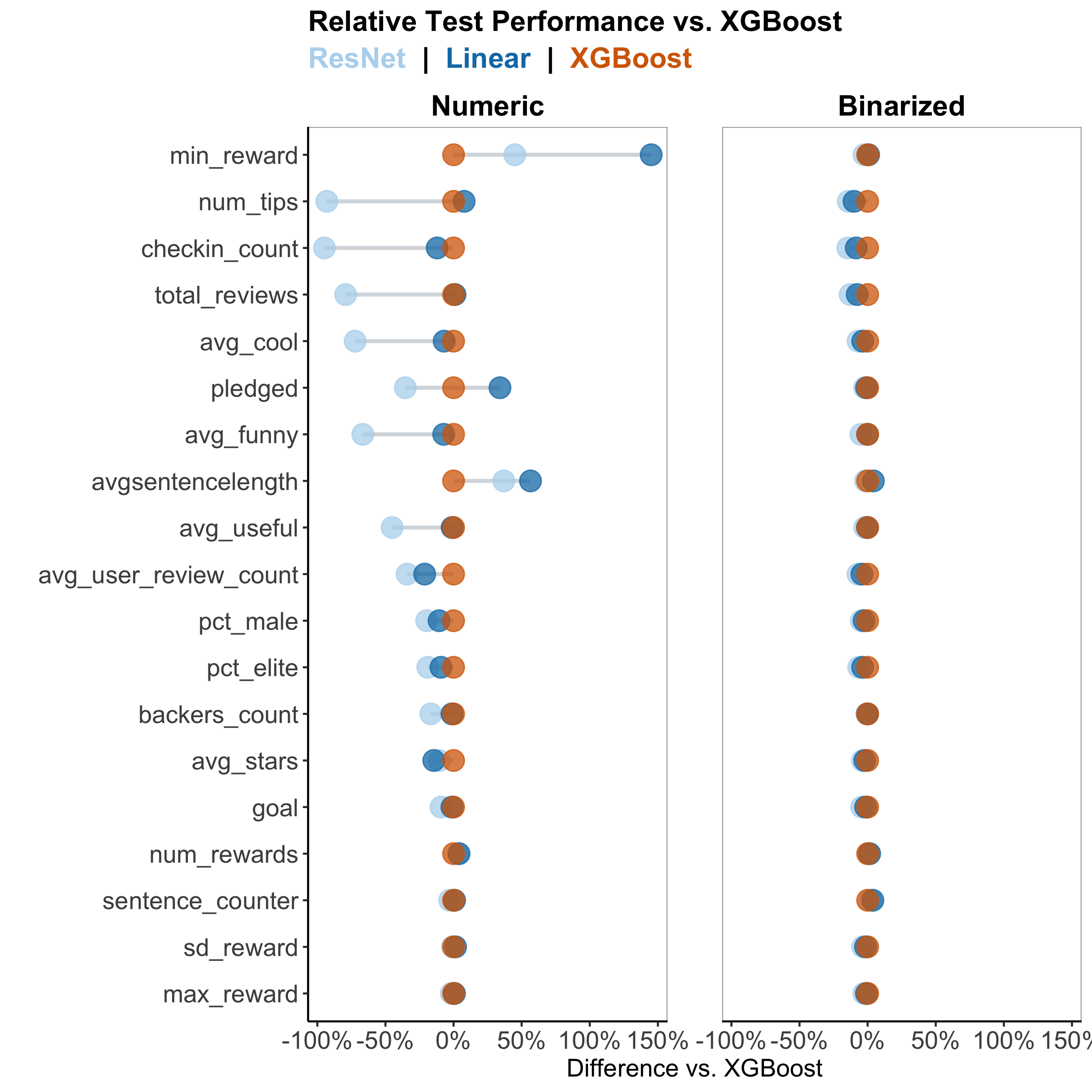}    
	\caption{Test performance of linear and ResNet models relative to XGBoost. \label{fig:model_selection}}
    \end{flushright}
\end{minipage}
\end{figure}

Figure \ref{fig:sampler} compares the test performance across two grid search sampling algorithms: simple random sampling, and Tree-Structured Parzen Estimator (TPE), a Bayesian sampler which has been found to yield improved results \citep{turner2021bayesian}. As before, the difference between the two samplers are much smaller for classification than for regression. We also note that for regression, the choice of which sampler performs better would depend on the HPO budget. 

\subsubsection{Model Selection}

All of the prior results in this paper have used XGBoost as the model family; however, we find that model selection is also more impactful for regression than for classification, as can be seen in Figure \ref{fig:model_selection}.\footnote{Note that due to resource constraints, these results include only 19 of the target variables examined in the prior sections.} This chart compares the average best tuned test performance of Linear and ResNet models relative to XGBoost. The differences between the best and worst-performing model families for classification are significantly less than those associated with regression. The average percent improvement for regression tasks between the worst and best-performing model family is 245.79$\%$. For classification, the average percent improvement is 6.34$\%$. The paired differences are statistically significant for $\alpha = 0.90$.\footnote{The core results for this paper have also been replicated for Linear and ResNet models and are included in the supplemental material.}

Taken along with the results in Section \ref{sect:sampler}, the implication is that heuristic choices in all parts of the ML pipeline matter relatively more for regression than for classification. The selection of models included in model selection are more consequential. Model selection can be significantly shortcut for classification because the best model is closer in performance to the worst and/or default model. With implications for replicability, the models chosen to benchmark performance in research proposing a new algorithm for regression also take on increased importance. These results also present an interesting tradeoff for researchers. Although regression requires more resources, there are also potentially larger benefits to be found when developing new regression algorithms (for all parts of the ML pipeline).

\section{DISCUSSION}

Our results bring additional nuance to the current understanding of the importance of HPO in the machine learning pipeline, particularly as it pertains to replicability of ML findings, sustainable AI, and automation of ML heuristics. HPO is necessary to achieve optimal performance in ML models \citep{bischl2023hyperparameter}, to the point where ML benchmarking results can be reversed depending on the extent of HPO conducted \citep{bouthillier2021accounting, dodge2019show}. This has contributed to a lack of replicability in the ML literature and calls for increased detail in reporting of experimental parameters \citep{dodge2019show}. HPO budget, i.e. number of search iterations or total time, is also a framework that has been used for evaluating the differences between induction algorithms; for instance, deep learning methods have been found to achieve comparable performance to tree-based ensemble methods on tabular data, but deep learning methods require far more computational resources \citep{gorishniy2021revisiting}. Our results leverage HPO budget as a dimension by which to compare the relative resources required by regression and classifications and reveal the large discrepancy in computational requirements between the two tasks. Our focus on HPO also highlights the fact that regression is more sensitive to both heuristic choices and randomness. This both makes regression modeling findings around regression harder to replicate and calls for larger grids (and even more computation) to be used in such contexts.

A major cost associated with HPO is the computation time that it requires, especially in the modern age of large language models and neural architecture search (NAS) \citep{strubell2019energy}. A full grid search trains and evaluates models using all possible hyperparameter combinations, although randomized grid search and its variations have been shown to be just as effective but much faster \citep{bergstra2012random}. Still, given the criticality of conducting a thorough grid search, HPO uses a tremendous amount of resources. These resource requirements leads to egregious quantities of carbon emissions \citep{strubell2019energy, schwartz2020green} and also inequities in who is able to contribute to the ML field \citep{strubell2019energy}. Our findings contribute to recent calls for more efficient ML algorithms \citep{strubell2019energy, schwartz2020green, dodge2019show} by improving the efficiency of the ML pipeline rather than any specific modeling algorithm: assuming that regression and classification are interchangeable from the perspective of performance in a downstream application, we show that classification requires a smaller grid search and fewer resources in general. 

The other cost of HPO is one that is common to the entire ML pipeline. There are numerous heuristic choices involved, such as which induction algorithms to try for comparison or optimization, which features to use, how much training data to acquire, how to set the HPO budget, which hyperparameters to tune, the size of the grid, and more. These choices are usually made by knowledgeable data scientists, who are in short supply \citep{he2021automl}. AutoML attempts to automate some of these choices, thereby streamlining the number of heuristic choices in the pipeline \citep{he2021automl}. For instance, recent work has focused on determining which hyperparameters for each common model family are tunable (i.e. where HPO effort is best spent) \citep{probst2019tunability}. This paper makes a fundamental contribution to the AutoML literature by instead evaluating tunability based on an underlying characteristic of the data being modeled: the formulation of the target variable. We find that regression tasks are overall more tunable, which has previously observed but not systematically evaluated \citep{sipper2022high}.

This work makes the significant assumption that regression and classification can be used interchangeably in some contexts and studies the effect of this choice on the resources required by the ML pipeline.\footnote{Of course, there are also situations where either formulation could be reasonably used but the downstream outcomes will differ; we leave a thorough exploration of the choice between regression and classification in terms of outcomes to future work. } Thus, it provides insight into the choice of whether or not to binarize by conducting a systematic comparison. Although past work in statistics has demonstrated that binarization leads to issues in traditional analyses, it frequently occurs in applied ML. We demonstrate that regression tasks are particularly costly in terms of required modeling effort; they require a higher HPO budget and greater amounts of training data, and the model selection process is less generalizable. Classification should be chosen when possible for the sake of efficiency, and smaller grids can be used. On the other hand, regression may present a greater opportunity for researchers who wish to publish impactful results; however, sufficiently large grid search, ensembling, and repeated sampling should be used to ensure replicability.

There are a few other apparent limitations in this work. First, most of our results use XGBoost to demonstrate the salient differences between the two tasks. We assert that using XGBoost may actually yield conservative results based on preliminary experiments with ilnear models and ResNet deep learning models. Second, our datasets are relatively small compared to the data typically used for truly computationally burdensome ML tasks. Once again, we believe that the performance differences between regression and classification seen in our results may be conservatively estimated compared to what would be seen with larger datasets, both in number of instances and number of features, based on the results in Sections \ref{sect:learningcurves} and \ref{sect:pipeline}. Finally we also note that tabular datasets of medium size are quite common in business applications. Future work could verify our findings with larger datasets and other model types.

%
\section{CONCLUSION}
We have experimentally compared the effect of choosing numeric regression vs. binary classification on the required resources and resulting performance in the ML pipeline. We show that choosing a numeric target variable consistently requires more time, computation, and data resources, and yields results that are more sensitive to randomness and model selection. We present actionable recommendations for ML researchers, users, and consumers of models.

\bibliography{tve_bib} \bibliographystyle{plainnat}





\end{document}